# Fotogrametri Verilerinde Yapım Bilgisinin Modellenmesine Yönelik Gömülü Biçim Eşlemesi

Demircan Taş[1]; Mine Özkar[2]
[1,2]İstanbul Teknik Üniversitesi
[1]http://demircantas.com, [1]demircan.tas@itu.edu.tr; [2]http://akademi.itu.edu.tr/ozkar/, [2]ozkar@itu.edu.tr

## Özet

*Mevcut yapıların fotogrametri yoluyla elde edilmiş üç boyutlu modellerinde, gözün seçebildiği tüm biçimler modelin geometrik bileşenlerinde her zaman karşılıklarını bulamaz. Oysa anlamlı parça ve bütünlerin hızlı ve detaylı belgeleme yöntemleriyle alınan kayıtlarla eşlenmesi, mevcut yapıların bilgi modellerinin oluşturulmasına yönelik avantaj sağlayacaktır. Bu probleme cevap üretmeyi hedeflerken, üç boyutlu modellerde örüntü tanımının zorluklarını aşmak için izdüşümle elde edilmiş iki boyutlu sade bir örneklem kullanılan çalışmamızda, fotogrametrik veriden yapı bilgi modeline doğru bir iş akışında kullanılabilecek bir biçim eşlemesi yöntemi geliştirilmiştir. Ortam ışığı kısıtlaması (ambient occlusion) gibi işleme teknikleri, eğrilik (curvature) ve normal haritaları, üç boyutlu yüzey niteliklerinin iki boyutlu veri setlerinde temsilini mümkün kılan, çağdaş bilgisayar grafiği uygulamalarında sıklıkla kullanılan öğelerdir. Önerdiğimiz yöntem, ışık tabanlı görselleştirme yerine, bu haritalamalar üzerinden örüntü tanınmasını temel alır. Uygulamanın ilk aşamasını tarihi mozaiklerin fotogrametri yoluyla taranması ve tarihi tuğla duvarların belgelere dayalı olarak üç boyutlu modellenmesi, ikinci aşamasını bu verilerden Alice Vision, OpenCV-Python, ve Autodesk Maya kullanılarak elde edilen yüzey temsilinin işlenmesi ile duvarların yapım bilgisini ve olası öz niteliklerini içeren modellerin oluşturulmasını kapsamaktadır. Sonraki aşamada elde edilen eşleme verisinin kural temelli tasarım ve yapım süreçlerindeki bilgiyi beslemesi öngörülmektedir.*

**Anahtar Kelimeler:** *Bilgisayarla görme, dijital kültürel miras, görme ve yapma, görsel hesaplama, üretken sistemler.*

# Embedded Shape Matching in Photogrammetry Data for Modeling Making Knowledge

## Abstract

*In three-dimensional models obtained by photogrammetry of existing structures, all of the shapes that the eye can select cannot always find their equivalents in the geometric components of the model. However, the matching of meaningful parts and assemblages with the records acquired with rapid and detailed documentation methods will provide an advantage for the creation of information models of existing structures. While aiming to produce answers to this problem and in order to overcome the difficulties of pattern recognition in three-dimensional models, we used two-dimensional samples obtained by projection. Processing techniques such as ambient occlusion, curvature and normal maps are commonly used in modern computer graphics applications that enable the representation of three-dimensional surface properties in two-dimensional data sets. The method we propose is based on the recognition of patterns through these mappings instead of the usual light-based visualization. The first stage of the application is photogrammetric capture of a few examples of Zeugma mosaics and three-dimensional digital modeling of a set of Seljuk era brick walls based on knowledge obtained through architectural history literature. The second stage covers the creation of digital models by processing the surface representation obtained from this data using Alice Vision, OpenCV-Python, and Autodesk Maya to include information on aspects of the making of the walls. What is envisioned for the next stages is that the mapping data contributes and supports the knowledge for rule-based design and making processes.*

## 1. Giriş

Mimarlık ve tasarım alanlarında yaygın kabul görmüş bazı çalışmalar, tasarım sürecini "görme ve yapma" (*seeing and doing*) ve "gör, hamle yap, gör" (*see-move-see*) gibi tanımlarla açıklarken, tasarım nesnesine müdahale etme ile nesnede öğeler arası yeni ilişkiler görmeyi bağlantılı olarak ele almıştır (Stiny, 2006; Schön ve Wiggins, 1992). Sürecin temel bir öğesi olan görme, bir bütünün içinden, gömülü (*embedded*) bir parça veya bütünün seçilmesi iken, diğer bir öğesi olan yapma, seçilmiş bir parça veya bütünü yenileriyle değiştirme veya bu işlemi özyinelemeli uygulama yoluyla gerçekleşir. Hamle yaptıkça, tasarımın nesnesi ve bağlamı dönüşerek, yeni görülere yol açar. Tasarımı, sürece ve tasarımın sonlu veya sonsuz çözümlenmiş öğelerine odaklı ele alan bir hesaplamalı tasarım yaklaşımı, Schön'ün (1983, 75) görme ve yapma ile döngü içinde ortaya konulabilecek önermeler için gerekli gördüğü sanal ortamı sunar. Özellikle üretken sistemler ile yeni tasarımların oluşturulmasında benimsenen bu yaklaşım, aynı zamanda tarih içinde tamamlanmış tasarım ürünlerinin geçmiş süreçlerinin incelenmesinde ve belgelenmesinde de kullanılmıştır (Knight, 1994). Tüm bu hesaplamalı tasarım uygulamaları için ihtiyaç duyulan sanal ortamın, eskiz ve maketlere ek olarak dijital temsillerinin oluşturulması da gelişen teknolojilerle beraber yaygınlaşmaktadır. Ancak bu ortam, mevcut kullanımlarda var olanın belirli değişkenler üzerinden modellenmesinden oluşmaktadır ve yukarıda vurgulanan yapma bileşeniyle ilgili olan malzeme ve yapım süreçlerinin tanımları geri planda kalmaktadır.

Var olan yapıların ve yapılı çevrenin belgelenmesinde sahada durum tespiti için kullanılan fotogrametri, LIDAR, panoramik video gibi çeşitli görüntüleme teknikleri ve bunlara dayanan sayısal modellerle gerçeğe yakın benzetimlerin oluşturulması en sık kullanılan dijital araç ve yöntemlerdir (Nagakura vd., 2015; 2017). Dijital görüntüleme teknikleri, çok çeşitli kullanımlara yönelik yüksek çözünürlüklü ve detaylı veri üretirken, bu verilerin yorumlanarak tasarım veya koruma amaçlı kullanımı tarihsel bakış açısı ile birlikte tasarım uzmanlığı gerektiren bir konudur. Var olan yapılara dair mimar bakış açısıyla oluşturulan görsel hesaplamalı betimlemelerde ise, tasarımın bileşenlerinin biçimsel olarak birbiriyle nasıl ilişkilendiği parametreler tanımlanarak kaydedilirken (Muller vd., 2006), mevcut literatürde biçime etki eden malzeme bilgisinin etkileri açıkça gösterilmemiştir. Güncel paradigmada, dijital temsilin malzeme ve yapım niteliklerini de içermesi, sıkça aranan bir özelliktir. Dijital modellerin hızla elde edilmesi için fotogrametri teknikleri kullanılabilmektedir. Mevcut yapılardan fotogrametri yöntemiyle elde edilen üç boyutlu temsiller, yüksek detay seviyesine sahip olmakla birlikte, parça bütün ilişkileri ve öznitelik verisi içermezler. Bu modellerin korumaya yönelik tasarım ve analiz süreçlerinde kullanılabilmesi için çoğu durumda fotogrametri yüzeylerinin kılavuz olarak kullanılarak yeni yüzeylerin modellenmesi gerekmektedir.

Çalışmalarımız, üretken tasarım ve yapım kurallarının uygulanabilmesi için, üç boyutlu ham temsiller üzerinde görme aşamasını modelleyen bir hesaplamalı araç üreterek, parça bütün ilişkilerinin, özellikle hasarlı yapıların veya tarama sürecinde veri kaybı oluşmuş veri setlerinin analiz edilebilmesini, eksik verinin veya parçaların ortaya çıkartılan üst düzey örüntüler doğrultusunda tamamlanabilmesini hedeflemektedir. Bu bildiri kapsamında sunulan aşamalarda, üç boyutlu modellerde örüntü tanımadan önce, izdüşümle elde edilmiş iki boyutlu ve dik açılı geometriler içeren bir örneklem kullanılarak, fotogrametrik veriden yapı bilgi modeline doğru gerçekleşen bir iş akışında geçerli olabilecek bir biçim eşlemesi yöntemi geliştirilmiştir.

## 2. Yöntem
### 2.1. Örneklem ve Belgeleme
Çalışma özünde fotogrametri gibi sahada hızlı ve güvenilir görüntü alma teknikleriyle elde edilen verilerin işlenmesine odaklanmaktadır. Ancak çok sayıda benzer model üzerinde çalışılması gerekirken, birden çok saha çalışması bu bildiri kapsamında mümkün olmamıştır. Var olan çevrimiçi fotogrametrik veriler olası işleme için incelenmiş, ancak verilerin zenginliği, yüksek işlem gücü gerektirdiğinden ve detayların fazla olmasından ötürü, belgeleme aşaması için basit örnekler üzerinde çalışılması uygun bulunmuştur.

Çalışma için belirlenen kapsam Orta Çağ Anadolusu'nda tuğla duvar örüntüleri olarak belirlenmiş, ön çalışma olarak tuğla dizimine benzer ama geometrik dizilimde çok daha sade düzenler sergileyen mozaiklere bakılmıştır. Zeugma Mozaik Müzesi'nde sergilenen 4 adet mozaik fotogrametrik yöntemlerle görüntülenmiştir. Her mozaik için yaklaşık 20 adet fotoğraf çekilmiştir.

Fotogrametri çalışmalarında sahada alınan görüntülerin işlenmesinde, ayrı ayrı alınan görüntülerin bir araya getirilmesi için görüntülerin düzeltilmesi gerekmektedir. Zeugma Mozaik Müzesi'nde fotogrametri için kaydedilen fotoğraflarda, ideal ışık ve konum koşullarının değerlendirilmesi yapılabilmiş, çekim açısı, mesafesi gibi kriterler belirlenmiştir. İdeal olmayan koşullar nedeniyle çekilen fotoğrafların büyük kısmının işlenmesi gerekmiştir.

Fotogrametri aşamasında GUI olarak Meshroom çalıştıran açık kaynaklı bir yazılım olan Alice Vision ve ticari sürümlü bir fotogrametri yazılımı olan Agisoft Metashape kullanılmıştır. Yazılımlara kullanıcı bazı düzeltmeler yapmak için müdahale edebilmektedir. Kaynak olarak kullanılan fotoğraflar üzerinden özellik tanıma (feature matching) ile eşlenen öğelerin koordinatlarıyla seyrek bir nokta bulutu oluşturulmuştur. Seyrek nokta bulutundan faydalanarak fotoğrafların üç boyutlu ortamda konum ve yön bilgileri elde edilmiştir. Konumlandırılmış fotoğrafların piksel verisi ile renk bilgisini de içeren yoğun bir nokta bulutu üretilmiştir. Yazılımda daha sonra nokta bulutu üçgen örgüye (triangular mesh) dönüştürülürken, UV koordinatlarına göre renk bilgisi taşınarak, gerçeğe benzer bir görüntü bütünleşik bir yüzey olarak elde edilmiştir (**Şekil 1**).

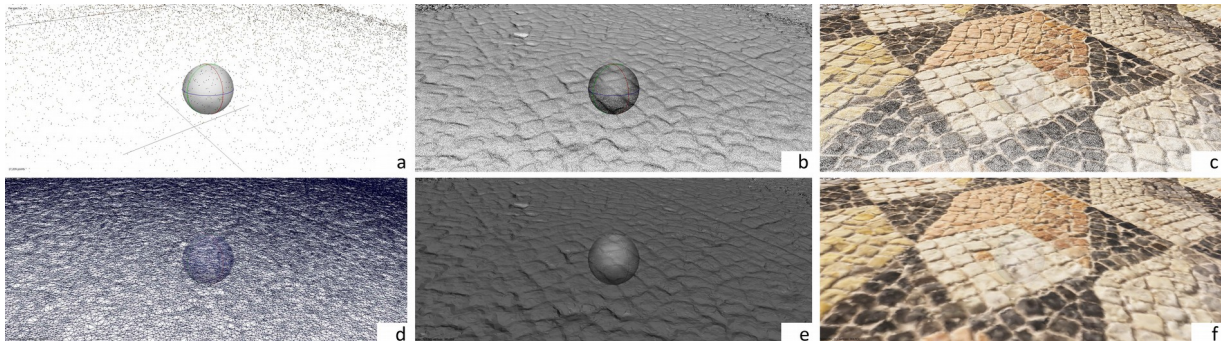

**Şekil 1:** Zeugma Mozaik Müzesi'nden bir örnek ve fotogrametri adımları: soldan sağa üst satır, seyrek nokta bulutu (a), yoğun nokta bulutu (b), renkli nokta bulutu (c), soldan sağa alt satır üçgen örgüyü (d), poligon yüzey (e) ve UV kaplamalı yüzeyi (f) göstermektedir.

Seçilen örneklem olan tarihi tuğla örgüleri için ise, çok sayıda örnek elde etmek için birçok ili kapsayacak saha çalışması bildiri dahilinde gerçekleştirilmemiş, örnekler mimarlık tarihi literatüründe var olan tespitlere dayandırılarak modellenmesiyle üretilmiştir. Bu modellerin işlenmesiyle yüzey bilgisi oluşturularak fotogrametrik verilere denk veri setleri oluşturulmuştur.

Fiziksel üretim ve malzeme, güncel hesaplamalı tasarım paradigmalarında yer edinmiştir. Bu sebeple çalışma, tasarımcının doğrudan müdahalesiyle üretilen soyutlamalar yerine, fiziksel üretimler veya mevcut yapıların taranmasıyla elde edilen modeller üstüne yoğunlaşmıştır. Yöntemin uygulanması ve başarısının tarihi tuğla duvar örneği üzerinden gösterilmesi planlanmıştır. Tarihi tuğla duvar örnekleri, dizilim tasarımında çeşitlilik, birimlerin tanımlı olması, geometrilerin kolay okunabilmesi, fotogrametriye uygun sade yüzey nitelikleri ve literatürde yapım bilgisini içeren çalışmalar (Bakırer, 1981) olması nedeniyle seçilmiştir. Oluşturulan yöntemin, tasarım süreci için gerekli akıcılığa sahip olmasına öncelik verilmiştir. Bu aşamada tuğla duvar için kullanılan örnekler Bakırer'in (1981) çalışmasından seçilmiştir. Örneklerin seçiminde, az sayıda birimle elde edilebilen görsel çeşitliliğe öncelik verilmiştir. İlk aşamada sadece birbirine dik tuğla içeren örnekler ele alınmıştır. Bunlar Harput Ulu Cami minare kaidesi ve gövdesi, Selimeköy'de anonim bir türbe, Akşehir Ulu cami minare pabucu, Malatya Ulu Cami tromplar, Konya Sahip Ata Hanikah, Erzurum Tepsi Minare gövdesi, Malatya Ulu Cami'de kubbeye geçişteki sağır nişler ve Pınarbaşı Melik Gazi Türbesi üzerindeki örüntülerdir (Bakırer 1981, ş. 44, 48-53, 59-66).

Tuğla örgü örnekleri modellenirken, Bakırer'in çizimlerinin aslına uygun olduğu varsayılmış ve modeller çizimlere sadık olarak ideal olmayan geometrilerle modellenmiştir. Örneğin, derz veya tuğlaların boyutları arasındaki farklılıklar çizim veya baskıdan olduğu kadar, tuğla üretimi ve diziminden kaynaklanabileceği düşünülmüştür. Modellerin üç boyutlu modelleme teknikleri kullanılarak kesişen poligonlardan oluşturulmuş, tuğlaları temsil eden poligonlar kopyalanırken çizimlerdeki bilgilere sadık kalınarak değişiklikler yapılmıştır. Konya İli'ndeki örnekler özelinde yapı malzemesinin değişken boyut bilgileri için Aktaş Yasa'nın (2016) çalışmasındaki detaylı veriler dikkate alınmıştır.

## 2.2. Görüntü İşleme ve Gömülü Biçim Eşleme

Bir sonraki aşama, fotogrametri çalışmasından ve üç boyutlu modellerden elde edilen iki boyutlu yüzeylerin işlenmesi olmuştur. Bu çalışmanın özgün yanı biçim tanıma algoritmalarını doğrudan üç boyutlu veri üzerinde çalıştırmak yerine, modelin izdüşümlerinden oluşturulan haritaların kullanımıdır. İki boyutlu haritaların oluşturulma sürecinde ışık bilgisine alternatif olarak yükselti, ortam ışığı kısıtlaması (*ambient occlusion*), yüzey normali ve eğrilik (*curvature*) haritaları kullanılmıştır. Fotogrametri için *Alice Vision*, modellerden iki boyutlu haritalamaların elde edilmesi için *Autodesk Maya* kullanılmıştır (Palamar, 2014).

Algoritmaların iki boyutlu izdüşümler üzerinde kurgulanması hem sürecin daha akıcı olmasını sağlaması, hem de insanların üç boyutlu ortam ve nesneleri algılayış yöntemlerine benzerliği sebebiyle tercih edilmiştir. Ayrıca mevcut biçim tanıma kütüphanelerinin çoğu, iki boyutlu veri kümeleri üzerinde hızlı sonuç vermektedir.

Ortam ışığı kısıtlaması, yüzey normali ve eğrisellik haritaları, iki boyutlu veri kümesinde üç boyutlu yüzey niteliklerini tanımlamayı mümkün kılmaktadır. Ayrıca bu haritalar, mevcut yapıların fotogrametrik modellerinden üretilebilmektedir. Bu haritalar üzerinde OpenCV-Python kütüphanesi etkinleştirilerek, hızlı biçim eşlemeleri elde edilebilmekte, eşlemelerden elde edilen koordinat ve öznitelik bilgisi üç boyutlu ortamda modelin zenginleştirilmesi için kullanılabilmektedir.

Biçim eşleme aşamasında yükselti haritalama ile elde edilen görüntülerde gömülü biçimlerin tanınması için, eğitilebilen tanıma algoritması HAAR kullanılmıştır. Şablon eşleştirme yöntemi ideal bir tipe dayalıdır. Sanal alanda kusursuz bir tuğla oluşturulabilir ve kusurlu benzerlerinin tanınması için bir model görevi görür. HAAR basamaklarının kullanımı için tek bir kaynak modeli yeterli değildir ve yaklaşım popülasyon düşüncesine benzer. Tanıma için genlere benzer bir özellik kütüphanesi tanımlamak için binlerce örnek gereklidir. Ortaya çıkan sistem daha sağlam ve daha kaliteli bir soyutlamanın temeli olabilir. Eşleme için kullanılacak şekiller başlangıç şeklinden ayrı olarak 3 boyutlu ortamda üretilip, farklı dosyalara 2 boyutlu haritalamalarla tanımlanmıştır. Tuğla birimine denk olan iki boyutlu bu şekiller tuğlanın farklı rotasyon ve boyutlarını da kapsamaktadır. Farklı görüntülerde bu şekillerin tanınması hedeflenmiştir. HAAR, eğitimle oluşturulmuş bir veri seti üzerinde çalışan bir algoritmadır. Eğitim için yüksek bir sayıda (binler) bir görüntünün türevi veya benzeri kullanılmaktadır. HAAR veri tabanları genelde yüz tanıma için kullanıldığından, açık kaynak olarak mevcut kütüphaneler çalışmamıza uygun olmamış, çalışma kapsamında görsel türetmemiz gerekmiştir. Üç boyutlu modellerden, üretken bir sistemle elde edilen görüntüler, boyutlar, olası hasar benzetimleri ve konum gibi değişkenlerle çoğaltılmıştır.

## 3. Model ve Bulgular

Çalışma yukarıda belirtildiği gibi iki veri seti üzerinden ilerlemiştir. Birinci veri seti, bilginin oluşturulması, ikinci veri seti bilginin işlenmesi üzerine yoğunlaşmıştır.

### 3.1. Mozaik Örneği

Zeugma Mozaik Müzesi'nden alınan bir örnek üzerinde belgelemeden işlenebilir görüntünün elde edilmesi aşamaları görülebilir. ZBrush'tan alınan normal haritasında kırmızı ve yeşil tonlar yatay ve düşey yönelime dair bilgi vermekte, mozaik taşlarının diziliminde üst düzenlere dair ipuçlarını barındırmaktadır. Örneğin Şekil 2'de, renklerle ayrışan prizma figürlerinin belli başlı hatlarının taşların dizilimindeki hizalardan da okunduğu görülmektedir.

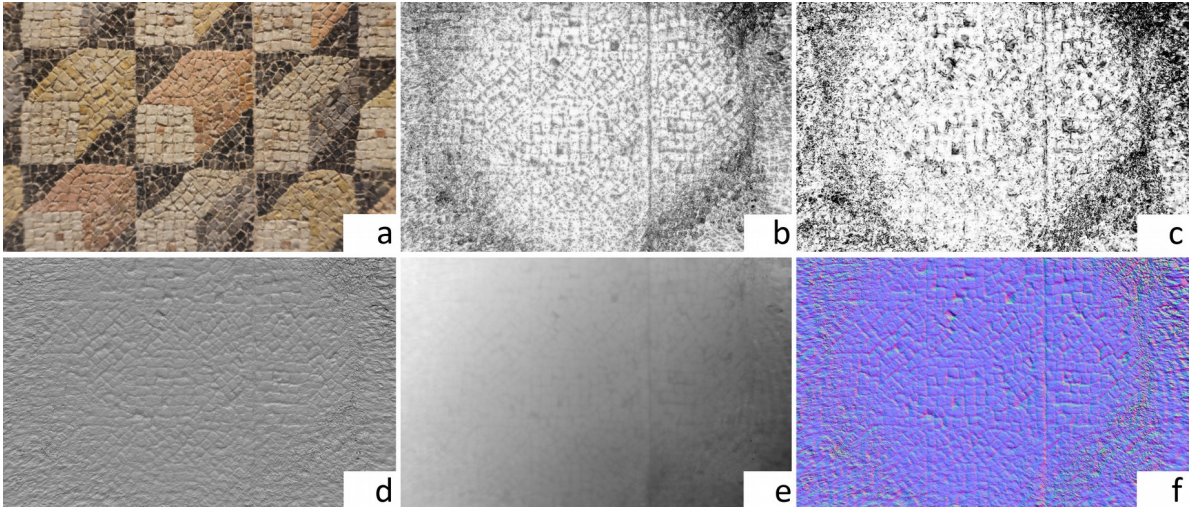

**Şekil 2:** Mozaiğin fotoğrafı ve 3 boyutlu modelden elde edilen haritalar: renk haritası (a), ortam ışığı kısıtlaması (b), eğrilik (c), renksiz yüzey(d), yükselti (e), normal(f).

Şekil 1-c,f'de sayısal modelin gözün kompozisyonun parçalarını ayırt etmesine olanak verdiğini ancak bu ayrıklıkların modelde bilgi olarak kaydedilmediğini görmekteyiz. Şekil 2-f'de ise oluşturduğumuz normal haritasında bu bilgiye sayısal olarak ulaşmayı kolaylaştıracak vurguları görebilmekteyiz. Tuğla duvarları ele alan ikinci aşama çalışmamızda, bunu aşmaya yönelik önerimiz ve bulgularımızı sunmaktayız.

### 3.2. Tuğla Örneği

Mevcut tarihi tuğla duvar örnekleri, teknik çizimler, ve literatürden edinilen verilerle üç boyutlu olarak modellenmiştir. Şekil 3'te Bakırer'in (1981) yayınında 44-c olarak belirtilen örneğe ait üç boyutlu modelleri görmekteyiz. Bu üç boyutlu modellerin oluşturulmasında, kesit bilgilerinin varlığı, daha sonra ön yüzeyin incelendiği aşamalarda ön yüzeydeki tuğla boyutlarından, tuğla derinliği ve tuğla tipi bilgilerinin de tespit edilebilmesi düşünülerek, önemsenmiştir. Bu örnek özelinde, sadece yatay olarak dizilen ve kuşağın üst ve alt sınırlarını çizen tuğlalar diğerlerine görece daha uzun ve derindir. Uzun tuğla olarak ön cephenin dijital yüzey modelinde tespit edilmesi, harç içindeki kesitte ne kadar derine saplandığına dair bilgiyi de çağıracaktır.

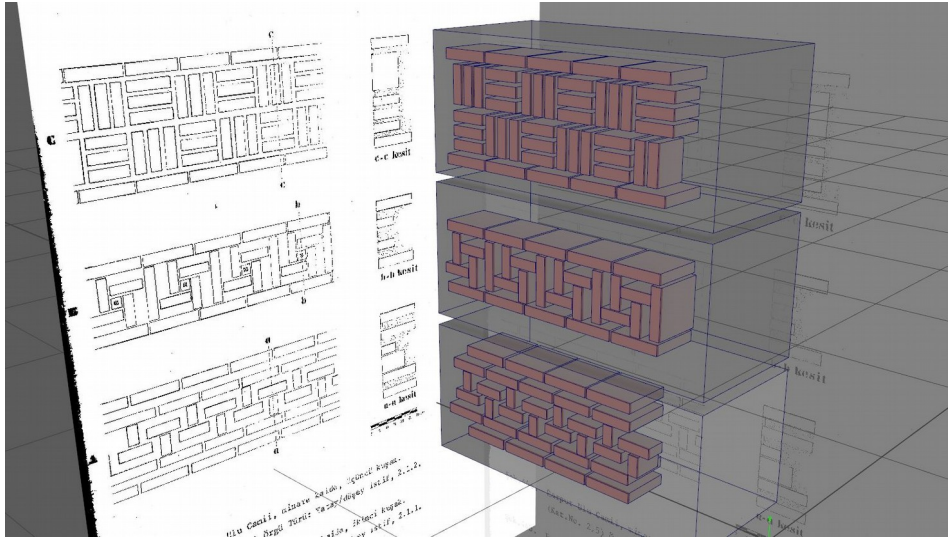

**Şekil 3:** Bakırer'in Harput Ulu Camii, minare kaidesindeki üç kuşak için oluşturduğu teknik çizimlere dayandırılarak elde edilmiş üç boyutlu model.

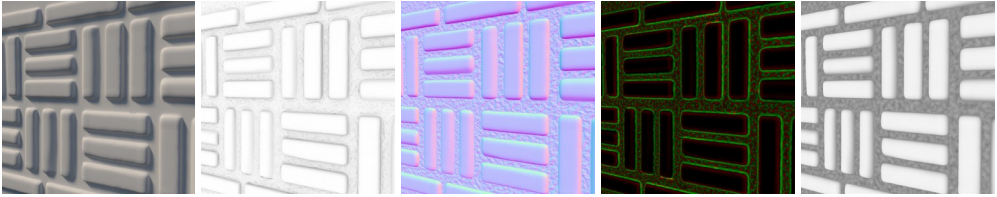

**Şekil 4:** Bakırer'in 44-c olarak tanımladığı üst kuşağa ait 3 boyutlu modelden, eşleme haritalarının elde edilişi. HAAR eğitimi ve eşleme süreci bu haritalarla gerçekleştirilmiştir.

Çalışmamızda HAAR kademelendirmesi (Viola & Jones, 2001) ile şekil eşlemesi, iki boyutlu yüzey modellerinde, belirtilen bir parçanın (tipik bir tuğlanın seçilmiş bir yüzeyi), verilen şekilde (duvar parçası) kaç adet olduğunun ve koordinatlarının tespitine indirgenmiş bir problemdir (Tablo 1). Çalışmanın bildiri kapsamında tamamlanan aşamalarında, eşleme için yüzeyde aranan şekil tek bir tuğlanın temsili olarak tanımlıdır. Farklı tuğla tiplerini ve tipler içinde boyutların yapım ve aşınmasından kaynaklı farklılaşmış durumlarını tanıyabilecek bir sistemin eğitimi amaçlanmıştır. Bu amaç doğrultusunda ideal bir tuğla yerine farklı ölçülere ve yüzeysel kusurlara sahip sonsuz sayıda tuğla oluşturabilecek bir üretken sistem tanımlanmış, zaman ve işlem gücü sınırlaması sebebiyle 400 örnek oluşturulmuştur. Oluşturulan örnekler ortam ışığı kısıtlaması, yüzey normalleri, eğrilik, ve yükselti haritaları oluşturmakta kullanılmıştır (Şekil 4).

**Tablo 1:** Eşleme için hazırlanan Python kodu

```
import cv2 # open computer vision kütüphanesi

brick_cascade = cv2.CascadeClassifier('Haar/cascade_brick.xml') # ürettiğimiz HAAR cascade veritabanı

img = cv2.imread('images/height1.png') # örüntülerin aranacağı görsel

while True:
```

```
    gray = cv2.cvtColor(img, cv2.COLOR_BGR2GRAY) # resim gri tonlara çevriliyor
    bricks = brick_cascade.detectMultiScale(gray, 10, 25) # argümanlar tarama aralığı
ve boyutunu sınırlıyor

    for (x, y, w, h) in bricks: # elde edilen uyum koordinatları için
        font = cv2.FONT_HERSHEY_SIMPLEX
        cv2.putText(img, 'brick', (x-w, y-h), font, 0.5, (0,127,255), 2, cv2.LINE_AA)
        cv2.rectangle(img, (x, y), (x+w, y+h), (255, 0, 0), 2) # metin ve dörtgen
oluşturma

    cv2.imshow('image', img) # sonucu gösterme komutu (gerçek zamanlı)

    k = cv2.waitKey(30) & 0xff
    if k == 27:
        break

cv2.destroyAllWindows()
```

Alternatif üst ilişkileri de göz önünde tutarak, ikili, üçlü gruplar gibi, örüntünün başka alt parçaları da aranan şekil olarak tanımlanabilmektedir. Çalışmada tek tuğla için bir kenar esas alınmıştır. Ancak birleşmeyen köşe tanımı da yapılabilmektedir.

Sonuçların başarısı ayarlara bağlı olarak değişmektedir (Şekil 5 ve Şekil 6). Denemeler sonucu, minimum yakın eşleme (*minNeighbors*) parametresinin eşleme kalitesine doğrudan etki ettiği saptanmıştır. Bu parametre, seçilen bir eşlemenin yakınında en az kaç olası eşlemeye sahip olması gerektiğini belirleyerek bir öğenin birçok defa tanınmasını önlemektedir. Sunulan sonuçlarda gösterilen farklılıklar bu parametrenin değiştirilmesiyle elde edilmiştir. Sonuç görseldeki her bir kare, bulunan eşlemenin etiket konumunu gösterir. Şekil 5'te *minNeighbors* değeri 25 olarak tanımlandığı için bir tuğla bir ila beş kere etiketlenmiştir. Şekil 6'da ise *minNeighbors* değeri 150 olarak tanımlanıp, her bir tuğlanın bir etiketle işaretlenmesine yol açmıştır. Değerin yüksek tanımlanması düşey tuğlaların tanınmasını zorlaştırmıştır. Bu durumda eğitim kütüphanesinin çoğunlukla yatay tuğla içermesinin de etkili olduğu söylenebilir.

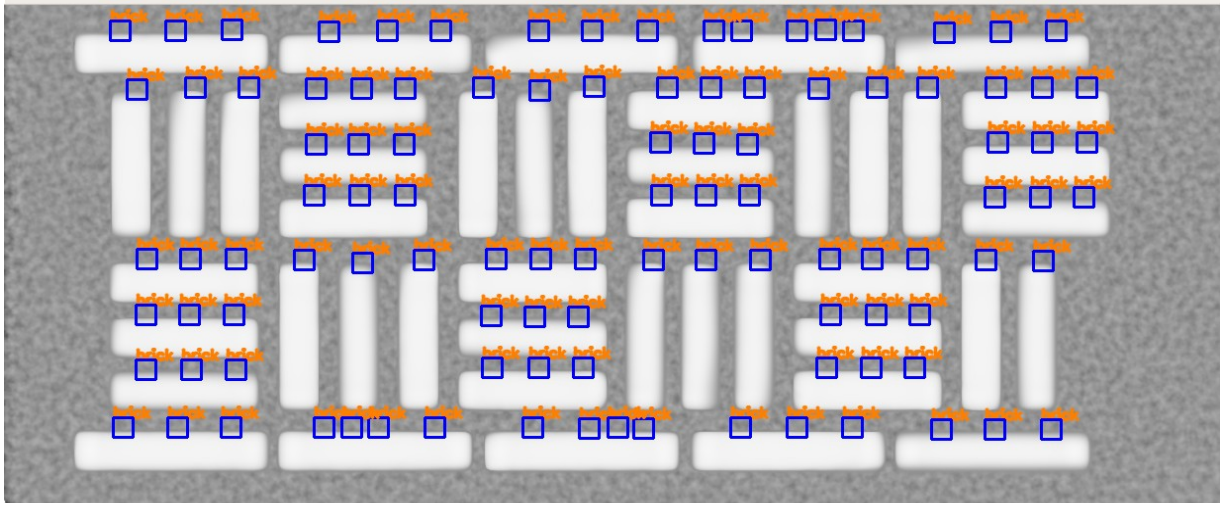

**Şekil 5:** Eşleme sonuçları 'detectMultiScale(25)'

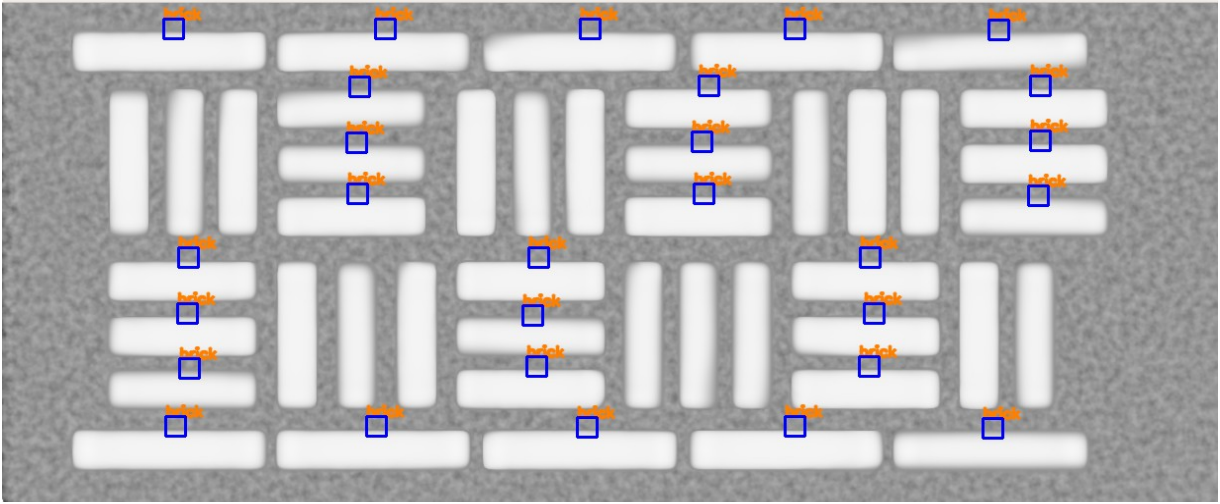

**Şekil 6:** Eşleme sonuçları 'detectMultiScale(150)'

## 4. Sonuçlar

Çalışmada biri tarihi mozaik, diğeri de tarihi tuğla duvar örüntüsü olmak üzere iki veri seti üzerinden görsel veri oluşturulması ve bunun işlenmesi üzerine yoğunlaşılmıştır. Çalışmada ilk olarak belgeleme ile ilgili tespitler yapılmıştır. Fotogrametrik yöntemlerle elde edilen görüntülerin nitelikleri büyük veri setlerinde mevcut yazılımlarla, küçük veri setlerinde kullanıcının doğrudan müdahaleleriyle düzeltilebilmektedir. Çok fazla sayısal verinin nitelikli olarak işlenmesi zaman ve işlem gücü gerektirmekte, yapım bilgisi bağlamında anlamlandırılmasına sıra gelememektedir. Örneklem sade tutularak bu sorun aşılmıştır. Eğitim verisi yetersiz olmakla birlikte, etiket konumları ve sayısı, doğru ayarların kullanımıyla isabetli hale getirilebilmiştir. Örneklemin oluşturulması araştırmada kendi başına bir tasarım süreci olarak ele alınmıştır. İdeal çalışmalarda örneklem sahada yapılan ölçüm ve belgeleme çalışmalarından edinilecektir.

Çalışmada önerilen yöntemin uygulanmasıyla mevcut fiziksel yapıların, öznitelik içeren soyutlamalarının hesaplamalı tasarım yöntemleriyle elde edilmesinde önemli bir adım atılmıştır. Tuğla duvar örneği özelinde, mevcut mimarlık kaynaklarından elde edilen teknik çizimlere göre oluşturulan üç boyutlu modeller, görsel belgeleme ve çözümleme süreçleri için kullanılmıştır. İleride, bir yandan, var olan bir yapıyı koruma veya geliştirme amacı güden tasarımcının, biçim kuralları üreterek çalışmasını, diğer yandan da yüzey temsillerinden yapım bilgisi içeren modellerin üretilmesini mümkün kılabilecektir. Bu yolla, tasarım ve yapı kullanım süreçleri sürerken, yeni kuralların eklenmesi sırasında yapılacak müdahalelerin hızla denenmesine imkan oluşturulabilecektir. Çalışmanın sonraki aşamasında elde edilen eşleme verisinin kural temelli tasarım eylemlerini beslemesine yönelik araştırmanın yapılması planlanmaktadır.

## KAYNAKLAR